\documentclass[runningheads]{llncs}
\usepackage{amsmath}
\usepackage{bbding}
\usepackage[T1]{fontenc}
\usepackage{graphicx}
\usepackage{amssymb}
\usepackage{multirow}
\usepackage{booktabs}
\usepackage{array}
\begin{document}
\title{Showing Many Labels in Multi-label Classification Models: An Empirical Study of Adversarial Examples}
\titlerunning{Showing Many Labels in Multi-label Classification Models}
%
\author{Yujiang Liu\inst{1} \and Wenjian Luo\inst{1,2} \inst{(}\Envelope\inst{)} \and Zhijian Chen\inst{1} \and Muhammad Luqman Naseem\inst{1}}
\authorrunning{Y. Liu et al.}
%
\institute{Guangdong Provincial Key Laboratory of Novel Security Intelligence Technologies, School of Computer Science and Technology, Harbin Institute of Technology, Shenzhen 518055, Guangdong, China \\
\email{23S151125@stu.hit.edu.cn} \\
\email{luowenjian@hit.edu.cn} \\
\email{\{21B951010,22bf51040\}@stu.hit.edu.cn} \and
Peng Cheng Laboratory, Shenzhen 518055, Guangdong, China \\}
\maketitle              
\begin{abstract}
With the rapid development of Deep Neural Networks (DNNs), they have been applied in numerous fields. However, research indicates that DNNs are susceptible to adversarial examples, and this is equally true in the multi-label domain. To further investigate multi-label adversarial examples, we introduce a novel type of attacks, termed "Showing Many Labels". The objective of this attack is to maximize the number of labels included in the classifier's prediction results. In our experiments, we select nine attack algorithms and evaluate their performance under "Showing Many Labels". Eight of the attack  algorithms were adapted from the multi-class environment to the multi-label environment, while the remaining one was specifically designed for the multi-label environment. We choose ML-LIW and ML-GCN as target models and train them on four popular multi-label datasets: VOC2007, VOC2012, NUS-WIDE, and COCO. We record the success rate of each algorithm when it shows the expected number of labels in eight different scenarios. Experimental results indicate that under the "Showing Many Labels", iterative attacks perform significantly better than one-step attacks. Moreover, it is possible to show all labels in the dataset.

\end{abstract}
\section{Introduction}
Deep Neural Networks (DNNs) have been widely applied in various domains. However, recent research has indicated that DNNs are susceptible to carefully crafted perturbations \cite{szegedy2014intriguing,dong2018boosting,wang2023towards}. These perturbations are imperceptible to the human eyes but can lead to incorrect classifications by DNNs. Such vulnerabilities can be particularly detrimental in safety-critical fields. For example, attackers could exploit them to compromise facial recognition systems \cite{hu2022protecting}, causing misidentification and potentially leading to financial loss for the victims. By studying adversarial examples, we can better understand their origins and devise appropriate defensive measures.

Currently, research on adversarial examples is primarily applied to multi-class classification tasks \cite{dabouei2019smoothfool,dong2022queryefficient,bai2023query}, where each sample is associated with a single label. In this scenario, attacks typically only need to consider the label with the highest confidence. Existing studies have demonstrated that existing methods can effectively achieve untargeted or targeted attacks in both white-box and black-box settings \cite{kurakin2017adversarial,chen2017zoo,xiao2018generating}. However, real-life samples often encompass multiple labels. For instance, a news article may cover topics such as politics, economy, and international affairs, while a landscape photograph may include elements like mountains, water, and white clouds \cite{boutell2004learning}. Therefore, studying multi-label adversarial examples holds greater practical significance.

The study of multi-label adversarial examples poses a greater challenge due to the interdependencies among labels. Specifically, when an attacker alters certain labels, it may lead to changes in other labels as well \cite{song2018multi}. This introduces new opportunities and challenges for both attackers and defenders. Several multi-label adversarial attacks \cite{song2018multi,zhou2020generating,kong2022evolutionary} have already been proposed and have been proven to effectively attack multi-label classifiers on popular multi-label datasets \cite{everingham2010pascal,chua2009nus,lin2014microsoft}. Recently, the performance of existing attack algorithms under different attacks has been studied \cite{zhou2021hiding,chen2024comprehensive}.

To further explore multi-label adversarial examples, this paper proposes a novel attack type, termed "Showing Many Labels". Our goal is to generate perturbations that result in as many positive labels as possible being included in the classifier's prediction results. To this end, we calculate the average number of positive labels per dataset and set the expected number of displayed labels at intervals of powers of two, then record the attack success rate. Based on the above considerations, we select eight attack algorithms that were adapted from the multi-class domain to the multi-label domain \cite{song2018multi,chen2024comprehensive}, as well as one attack algorithm specifically for multi-label domain \cite{zhou2020generating} for evaluation. Experimental results demonstrate that some algorithms can effectively show the majority of labels, and even all labels.

In summary, the main contributions of this paper are described as follows.
\begin{itemize}
	\item We introduce a new type of attacks, "Showing Many Labels", and evaluate the performance of nine attack algorithms under this attack. Among the eight adapted attack algorithms, FGSM \cite{goodfellow2015explaining}, FGM \cite{miyato2021adversarial}, DeepFool \cite{moosavidezfooli2016deepfool}, and C\&W \cite{carlini2017evaluating} were adapted by Song et al \cite{song2018multi}, while BIM \cite{kurakin2017adversarial}, PGD \cite{madry2019deep}, SLIDE \cite{tramèr2019adversarial}, and MI-FGSM \cite{dong2018boosting} were adapted by Chen et al \cite{chen2024comprehensive}. The MLA-LP was specifically designed for the multi-label environment by Zhou et al \cite{zhou2020generating}. We record the attack success rates of these algorithms under various conditions, providing a baseline for future research on multi-label adversarial examples.
    \item We assess nine attack algorithms on two models across four datasets. Experimental results indicate that iterative attack algorithms are capable of showing more labels compared to one-step attack methods. Furthermore, as the expected number of displayed labels increases, the attack becomes more challenging, with only a few attack algorithms being able to show all labels under specific models and dataset conditions.
\end{itemize}

The rest of this paper is organized as follows. Section \ref{section 2} introduces related work. Section \ref{section 3} provides a problem description. Section \ref{section 4} introduces the attack algorithms, models and datasets used in the experiments. Section \ref{section 5} analyzes experimental results. Section \ref{section 6} briefly summarizes this paper.

\section{Related Work}
\label{section 2}

\subsection{Multi-label Adversarial Attack}
In the multi-label learning domain, due to the interdependencies between labels, altering the predicted confidence of one label could lead to changes in the predicted confidence of other labels. Consequently, attack methods designed for multi-class classification tasks cannot be directly applied to multi-label scenarios. Currently, there is relatively little work on multi-label adversarial examples, with only a few attack methods available. 

Song et al. \cite{song2018multi} divided the multi-label learning problem into classification and ranking, and designed corresponding attack frameworks for each. Based on these two frameworks, they proposed four multi-label attacks: ML-DP, ML-CW, Rank I, and Rank II. ML-DP and ML-CW are used for attacking multi-label classification, while Rank I and Rank II are for multi-label ranking. Zhou et al. \cite{zhou2020generating} defined an optimization problem similar to ML-DP and then transformed the optimization problem into a linear programming problem. This method controls the size of the perturbation by limiting the $l_{\infty}$ norm of the perturbation. Hu et al. \cite{hu2021tkml} proposed an attack algorithm for the Top-k multi-label learning problem, $T_kML-AP$. This algorithm designed a new loss function, achieving untargeted attacks, universal untargeted attacks, and targeted attacks. Kong et al. \cite{kong2022evolutionary} proposed a black-box attack for generating multi-label adversarial examples based on the differential evolution algorithm. They designed a complementary mutation operator in the algorithm, enhancing the overall performance of the algorithm.

\subsection{Existing Attack Types}
To evaluate the effectiveness of attack algorithms, Song et al. \cite{song2018multi} proposed five types of attacks, which are "Hiding single", "Random", "Extreme", "Reduction", and "Augmentation". In the "Hiding single", adversaries select samples that contain at least two labels and randomly choose one label to hide. Adversarial examples generated ensure that the classifier is unable to identify the hidden label. In the "Random", adversaries randomly select a positive label and a negative label within a sample, and then launch attacks to alter the positive label to be negative and the negative label to be positive. "Extreme" is the most radical form of attack, aiming to invert all labels within a sample—turning positive labels into negative and vice versa. "Reduction" is an attack where attackers select a specific label and samples that contain at least two labels, one of which is the chosen label, with the goal of preventing the classifier from identifying the chosen label within the sample. "Augmentation" is the converse of "Reduction", the goal of attackers is to manipulate the classifier’s prediction to include the specific label through the attack. Zhou et al. \cite{zhou2021hiding} proposed the "Hiding all" attack type, where the attacker's objective is to hide all labels, rendering the classifier completely ignorant of the environment.

In this paper, we propose a novel attack type termed "Showing Many Labels", with the objective of generating adversarial examples that maximize the number of positive labels in the classifier's prediction results, ideally showing all labels present in the dataset. By proposing new attack type, we aim to provide a new baseline for future research on multi-label adversarial examples.

\section{Problem Description}
\label{section 3}

Suppose we have a multi-label classification problem with $l$ labels, where the input data is represented as ($\mathbf{x}$, $\mathbf{y}$). $\mathbf{x} \in \mathbb{R}^d$ denotes the feature vector of an example, and $\mathbf{y} \in \{-1, 1\}^l$ represents its label vector. When $y_i=1$, it indicates that the example contains the label $i$; otherwise, $y_i=-1$.

We denote the multi-label classifier by $f:\mathbb{R}^d \rightarrow \mathbb{R}^l$, which outputs the predicted confidence for each label of the exmaple. $f$ can be considered as a composition of $l$ sub-functions, namely $f=\{f_1,...,f_l\}$, where $f_i(\mathbf{x})$ represents the predicted confidence of the label $i$. Suppose $t$ represents a threshold; if the predicted confidence $f_i(\mathbf{x})$ for label $i$ is not less than $t$, it implies that  label $i$ is present in the example, otherwise it does not. In this paper, unless specifically referred to, the threshold $t$ is set to 0, and $f_i(\mathbf{x}) \in [-1, 1]$. Therefore, for any label $i$, the predicted outcome $y_{i}^{\prime}$ of the multi-label classifier can be represented as follows:
\begin{equation}
   y_{i}^{\prime}=\left\{
	\begin{aligned}
	1, \quad if \,\, f_i(\mathbf{x}) \geq 0\\
        -1, \quad if \,\, f_i(\mathbf{x}) < 0\\
	\end{aligned}
	\right
	.
        \label{eq1}
\end{equation}
When each dimension of $\mathbf{y}$ and $\mathbf{y}^{\prime}$ is equal, the classification is correct.

In the task of generating multi-label adversarial examples, we denote the adversarial perturbation by $\mathbf{r}$. The adversarial example is represented as $\mathbf{x}^{\ast}=\mathbf{x}+\mathbf{r}$. We aim to achieve the attack target while minimizing the perturbation; thus, the problem can be formulated with the following equation:
\begin{equation}
    \begin{aligned}
        &\min_{\mathbf{r}} \quad \lVert \mathbf{r} \rVert \\
        &\text{s.t.} \quad F(f(\mathbf{x}^{\ast})) = \mathbf{y}^{\ast}
    \end{aligned}
    \label{eq2}
\end{equation}
where $\mathbf{y}^{\ast}$ is the attack target, $F(\cdot)$ is the function represented by Equation \eqref{eq1}. 

In the "Showing Many Labels", our goal is to manipulate the classifier's predictions such that the number of positive labels meets the expected number of displayed labels. We calculate the average number of positive labels for each dataset and use the following formula to compute a range of expected numbers of displayed labels:
\begin{equation}
    \label{eq3}
    \resizebox{\textwidth}{!}{
    $expLabels=\left\{
    \begin{aligned}
        \lfloor avgLa&bels+2^n \rfloor, \quad if \,\, 0 \leq n < \lceil \log_{2}{(avgLabels}+2^n) \rceil \\
        &l, \quad if \,\, n=\lceil \log_{2}{(avgLabels}+2^n) \rceil
    \end{aligned} 
    \right
    .$
    }
\end{equation}
Where $0 \leq n \leq \lceil \log_{2}{(l-avgLabels)} \rceil$ is an integer, $expLabels$ represents the expected number of labels to be displayed, and $avgLabels$ denotes the average number of positive labels in the current dataset. In this paper, we set $\mathbf{y}^{\ast}$ to $\{1\}^l$. An attack is deemed successful when the number of positive labels in $F(f(\mathbf{x}^{\ast}))$ reaches $expLabels$. 

\section{Experimental Design}
\label{section 4}

\subsection{Attack Algorithms}
This subsection introduces the nine attack algorithms used in the experiments. 

\textbf{1) FGSM:} The FGSM \cite{goodfellow2015explaining} is a white-box attack applied in multi-class environments. It calculates the gradient of the model output with respect to the input and adds a small perturbation in the direction of the gradient's sign to achieve the attack. Since this method generates adversarial samples in a single step, it is considered a one-step attack. In this paper, we utilize the version of FGSM adapted for the multi-label environment \cite{song2018multi}. Specifically, compute the loss $\mathcal{L}(\mathbf{y}^{\ast},f(\mathbf{x}))$ between the attack target $\mathbf{y}^{\ast}$ and the model's output $f(x)$. The gradient of the loss for the input $\mathbf{x}$ is $\nabla_\mathbf{x}\mathcal{L}(\mathbf{y}^{\ast},f(\mathbf{x}))$. To minimize the loss $\mathcal{L}(\mathbf{y}^{\ast},f(\mathbf{x}))$, the perturbation is added to the input in the direction of $-\nabla_\mathbf{x}\mathcal{L}(\mathbf{y}^{\ast},f(\mathbf{x}))$, thereby generating an adversarial example. The algorithm can be described as follow:
\begin{equation}
    \mathbf{x}^{\ast}=\mathbf{x}+\epsilon \cdot sign(-\nabla_\mathbf{x}\mathcal{L}(\mathbf{y}^{\ast},f(\mathbf{x})))
\end{equation}
where $\epsilon$ is the perturbation step size, $sign(\cdot)$ is the sign function, and $\mathbf{x}^{\ast}$ denotes the generated adversarial examples.

\textbf{2) FGM:} FGM \cite{miyato2021adversarial} is an extension of FGSM. Unlike FGSM, which simply adds the sign of the gradient to the input to create an adversarial example, FGM uses $L_2$ norm normalization to better preserves the direction opposite to the gradient, potentially leading to more effective adversarial examples. Song et al. \cite{song2018multi} adapted the FGM to the multi-label environment and the core idea can be described as follow \cite{song2018multi}:
\begin{equation}
    \mathbf{x}^{\ast}=clip\left\{\mathbf{x}-\epsilon \cdot \frac{\nabla_\mathbf{x}\mathcal{L}(\mathbf{y}^{\ast},f(\mathbf{x}))}{\lVert \nabla_\mathbf{x}\mathcal{L}(\mathbf{y}^{\ast},f(\mathbf{x})) \rVert_2}\right\}
\end{equation}
where $\lVert \cdot \rVert_2$ is $l_2$ norm.

\textbf{3) BIM:} FGSM and FGM are both one-step adversarial attack methods. Their simplicity often leads to failure when attacking large datasets. To enhance the efficacy of adversarial attacks, Kurakin et al. \cite{kurakin2017adversarial} proposed an iterative variant of FGSM known as BIM. The BIM algorithm iteratively generates adversarial examples by making small, incremental adjustments to the input. With each iteration, it moves closer to the model's decision boundary with a small step size. The perturbed example are then clipped to ensure they remain within the valid range, ultimately creating examples that can mislead the model. The core function can be expressed as follow \cite{chen2024comprehensive}:
\begin{equation}
    \mathbf{x}^{\ast}_{n+1}= clip\left\{\mathbf{x}^{\ast}_n-\epsilon \cdot sign(\nabla_{\mathbf{x}^{\ast}_{n}}\mathcal{L}(\mathbf{y}^{\ast},f(\mathbf{x}^{\ast}_{n})))\right\}
\end{equation}
where $\mathbf{x}^{\ast}_{n+1}$ denotes the adversarial example generated during the $(n+1)$-th iteration. At the commencement of the BIM, $\mathbf{x}^{\ast}_0=\mathbf{x}$.

\textbf{4) PGD:} PGD \cite{madry2019deep} shares similarities with the BIM, with a key distinction being that PGD employs projection to ensure the results fall within a valid range. Additionally, PGD incorporates random perturbations to the input as the initial value. In this paper, we utilize a variant of PGD tailored for multi-label scenarios, with the core algorithm presented as follows \cite{chen2024comprehensive}:
\begin{equation}
    \begin{aligned}
        &\mathbf{x}^{\ast}_0=\mathbf{x}+\mathbf{d} \\
        &\mathbf{x}^{\ast}_{n+1}= \Pi_{\mathbf{x}+\mathcal{S}}\left\{\mathbf{x}^{\ast}_n-\epsilon \cdot sign(\nabla_{\mathbf{x}^{\ast}_{n}}\mathcal{L}(\mathbf{y}^{\ast},f(\mathbf{x}^{\ast}_{n})))\right\}
    \end{aligned}
\end{equation}
where $\mathbf{d}$ represents the random perturbation. $\Pi$ denotes the projection operation, which constrains the results within the range $\mathbf{x}+\mathcal{S}$. $\mathcal{S}$ signifies the permissible perturbation space.

\textbf{5) SLIDE:} SLIDE \cite{chen2024comprehensive,tramèr2019adversarial} is an adversarial attack strategy that targets the $l_1$ norm. This attack generates adversarial examples by controlling the sparsity of gradients, which implies that the attack modifies only a small portion of the input at each update step, rather than updating all dimensions at every step as is traditionally done with PGD.

\textbf{6) MI-FGSM:} MI-FGSM \cite{dong2018boosting} is an extension of the BIM that effectively addresses the issue of becoming trapped in local optima by incorporating a momentum accumulation mechanism during the iterative process. This approach not only enhances the success rate of attacks in white-box environments but also significantly improves the efficacy of attacks in black-box settings. In this paper, we employs the MI-FGSM tailored for the multi-label domain \cite{chen2024comprehensive}, with the core algorithm presented as follows:
\begin{equation}
    \begin{aligned}
        &\mathbf{g}_0=0,\mathbf{x}^{\ast}_0=\mathbf{x}, \\
        &\mathbf{g}_{n+1}=\mu \cdot \mathbf{g}_{n}+\frac{\nabla_{\mathbf{x}^{\ast}_{n}}\mathcal{L}(\mathbf{y}^{\ast},f(\mathbf{x}^{\ast}_{n}))}{\lVert \nabla_{\mathbf{x}^{\ast}_{n}}\mathcal{L}(\mathbf{y}^{\ast},f(\mathbf{x}^{\ast}_{n})) \rVert_1} \\
        &\mathbf{x}^{\ast}_{n+1}=\mathbf{x}^{\ast}_{n}+\epsilon \cdot sign(\mathbf{g}_{n+1})
    \end{aligned}
\end{equation}
where $\mu$ is decay factor, $\mathbf{g}_n$ is the momentum factor at $n$-th iteration, $\lVert \cdot \rVert_1$ is $l_1$ norm.

\textbf{7) ML-DP:} ML-DP \cite{song2018multi} is a multi-label adversarial attack algorithm transplanted from DeepFool \cite{moosavidezfooli2016deepfool}. It assumes that the perturbation has a linear impact on the output, and transforms non-linear constraints into linear constraints through first-order Taylor expansion. The optimization problem that this method aims to solve is presented as follows:
\begin{equation}
    \begin{aligned}
        &\min_{\mathbf{r}} \quad \lVert \mathbf{r} \rVert_2 \\
        &\text{s.t.} \quad -\mathbf{y}^{\ast} \odot \left(f(\mathbf{x})+\frac{\partial f(\mathbf{x})}{\partial\mathbf{x}} \cdot \mathbf{r}\right) \leq 0
    \end{aligned}
\end{equation}
In this context, $f(\mathbf{x}) \in [-1, 1]$, and the threshold $t=0$. However, ML-DP does not directly solve the aforementioned optimization problem. Instead, it utilizes a greedy strategy to search for adversarial perturbations that satisfy the attack objectives. This can lead to results that are not the global optimum, and the perturbations obtained may be large.

\textbf{8) ML-CW:} ML-CW \cite{song2018multi} is an adversarial attack method for multi-label learning models, based on the approach proposed by C\&W \cite{carlini2017evaluating}. This method transforms the constraints in the optimization problem into regularizers, such as employing the hinge loss. It then utilizes gradient descent to minimize the loss function. The core algorithm of ML-CW is presented as follows:
\begin{equation}
    \mathcal{L}=\lVert \mathbf{r} \rVert + \lambda \sum_{i=1}^{l} \max(0, -\mathbf{y}_i^{\ast}  f_i(\mathbf{x+r}))
\end{equation}
Where, the parameter $\lambda$ is a regularization term that balances the magnitude of the perturbation and the success rate of the attack. The algorithm employs a binary search to determine the appropriate regularization parameter. Here, $f(\mathbf{x}) \in [-1, 1]$, and the threshold $t=0$.

\textbf{9) MLA-LP:} MLA-LP \cite{zhou2020generating} transforms the optimization problem into a linear programming problem and controls the magnitude of the perturbation by minimizing its $l_{\infty}$ norm. To minimize the changes in the confidence of non-target labels while significantly altering the confidence of the target labels, MLA-LP defines the following optimization problem.
\begin{equation}
    \begin{aligned}
        &\min_{\mathbf{r}} \quad \lVert \mathbf{r} \rVert_{\infty} \\
        &\text{s.t.} \quad \frac{\partial \mathcal{L}(\mathbf{y}^{\ast},f(\mathbf{x}))}{\partial \mathbf{x}} \cdot \mathbf{r} \leq \mathcal{L}(\mathbf{y}^{\ast},\mathbf{t})-\mathcal{L}(\mathbf{y}^{\ast},f(\mathbf{x}))
    \end{aligned}
    \label{eq10}
\end{equation}
Where $\mathbf{t} \in \mathbb{R}^l$ is a set of threshold. By introducing the variable $\mathbf{z} \geq |r_i|$, Equation \eqref{eq10} can be rewritten as follows:
\begin{equation}
    \begin{aligned}
        &\min_{\mathbf{z},\mathbf{r}} \quad \mathbf{z} \\
        &\text{s.t.} \quad \mathbf{z} \geq r_i, \quad i=1,2, \dots ,d \\
        &\hphantom{s.t\quad} \mathbf{z} \geq -r_i, \quad i=1,2, \dots ,d \\
        &\frac{\partial \mathcal{L}(\mathbf{y}^{\ast},f(\mathbf{x}))}{\partial \mathbf{x}} \cdot \mathbf{r} \leq \mathcal{L}(\mathbf{y}^{\ast},\mathbf{t})-\mathcal{L}(\mathbf{y}^{\ast},f(\mathbf{x}))
    \end{aligned}
\end{equation}

\subsection{Classification Models}

This subsection introduces two multi-label classification models. The ML-LIW model was proposed by Song et al. \cite{song2018multi}, while the ML-GCN model was proposed by Chen et al. \cite{chen2019multilabel}.

\textbf{1) ML-LIW:} Song et al. \cite{song2018multi} introduced ML-LIW, a multi-label classifier based on the Inception v3 network pre-trained on the ImageNet dataset. To accommodate the requirements of multi-label classification, Song et al. replaced the original softmax layer with a sigmoid layer. Furthermore, to enhance model performance, they designed a comprehensive loss function that combines instance-wise and label-wise losses. Specifically, the instance-wise loss employs a modified instance AUC score, which helps capture the relationships between labels within each instance. The label-wise loss, on the other hand, incorporates ranking loss and label-wise AUC scores, aiming to alleviate the issue of label imbalance in the dataset.

\textbf{2) ML-GCN:} Chen et al. \cite{chen2019multilabel} proposed a multi-label classification model based on Graph Convolutional Networks (GCNs), known as ML-GCN. This model simulates the dependencies between labels by constructing a directed graph, where each node represents an object label and is represented through word embeddings. The core concept of ML-GCN is to leverage GCNs to learn node representations on the label graph, thereby mapping to a set of interdependent classifiers. The model employs a novel reweighting scheme to construct the label correlation matrix, which effectively guides the propagation of information in the GCN by balancing the weights between nodes and their neighborhoods. This approach not only mitigates the issues of overfitting and over-smoothing, but also captures complex relationships between labels, thereby enhancing the accuracy of classification.

\subsection{Datasets}
This subsection introduces four commonly used datasets in the multi-label learning domain.

\textbf{1) VOC2007:} The VOC2007 \cite{everingham2010pascal} dataset is a widely used benchmark in the field of computer vision. Comprising images from real-world scenarios, it includes 20 distinct categories. The training set comprises 5,011 samples, the validation set 2,510, and the test set 4,952.

\textbf{2) VOC2012:} The VOC2012 \cite{everingham2010pascal} dataset inherits and expands upon the VOC2007 dataset, featuring 20 categories. It includes a training set with 5,017 images and a validation set with 5,821 images and no test set.

\textbf{3) NUS-WIDE:} The NUS-WIDE \cite{chua2009nus} dataset is a web image dataset with multiple labels that created by the Media Search Lab at the National University of Singapore. It comprises 269,648 images across 81 categories. However, some samples are currently inaccessible. We use the accessible samples as described in \cite{ridnik2021asymmetric}. The number of accessible samples amounts to 169,823, which includes a training set of 119,103 and a validation set of 50,720 and no test set.

\textbf{4) COCO:} The COCO \cite{lin2014microsoft} dataset supports a variety of computer vision tasks, including image classification, object detection, segmentation, and image captioning. It encompasses 80 categories, with a training set of 82,783 images, a validation set of 40,504 images, and a test set of 40,775 images.

\section{Experimental Results and Analyses}
\label{section 5}

\subsection{Experimental Setting}
Due to the absence of a test set in both the VOC2012 and NUS-WIDE datasets, we randomly select 20\% of the data from the training and validation sets to serve as the test set. We resize all images to 448*448 and normalized the pixel values to the range $[0, 1]$.

For ML-GCN, we directly utilize the open-source code from the original paper and train it on four datasets. The code is available at https://github.com/Megvii-Nanjing/ML-GCN. For ML-LIW, we replicate it based on \cite{song2018multi}, and similarly, train it on four datasets.

We evaluate the model's performance on the datasets using five metrics: hamming loss, ranking loss, micro-F1, macro-F1, and average precision. The results are shown in Table \ref{tab1}.
\begin{table*}
    \centering
    \caption{The performance of ML-GCN and ML-LIW on four datasets}
    \resizebox{\textwidth}{!}{
    \begin{tabular}{c|c|c|c|c|c|c}
        \toprule[2pt]
        Model & Dataset & Hamming Loss & Ranking Loss & Average Precision & Micro-F1 & Macro-F1\\
        \midrule
        \multirow{4}{*}{ML-GCN} & VOC2007 & 0.0217 &0.0095 & 0.9590 & 0.8554 & 0.8343 \\
        & VOC2012 & 0.0200 & 0.0090 & 0.9634 & 0.8646 & 0.8505 \\
        & NUS-WIDE & 0.0158 & 0.0139 & 0.8407 & 0.7286 & 0.5746 \\
        & COCO & 0.0146 & 0.0132 & 0.9062 & 0.7812 & 0.7466 \\
        \midrule
        \multirow{4}{*}{ML-LIW} & VOC2007 & 0.0370 & 0.0159 & 0.9344 & 0.7778 & 0.7866\\
        & VOC2012 & 0.0403 & 0.0153 & 0.9382 & 0.7622 & 0.7609 \\
        & NUS-WIDE & 0.0577 & 0.0264 & 0.7582 & 0.4913 & 0.35 \\
        & COCO & 0.0860 & 0.0276 & 0.8002 & 0.4422 & 0.4164 \\
        \bottomrule[2pt]
    \end{tabular}
    }
    \label{tab1}
\end{table*}

In this paper, we employ the AdverTorch framework to implement six adversarial attack algorithms: FGSM, FGM, BIM, MI-FGSM, PGD, and SLIDE. This framework, which is based on PyTorch, provides a toolkit for adversarial attacks and defenses. To ensure optimal performance of these algorithms, we meticulously set the parameters for each algorithm based on the experimental results provided in the original papers. For the FGSM algorithm, we set the parameter $\epsilon$ in Equation (3) to 0.3. In the FGM algorithm, we set the parameter $\epsilon$ in Equation (4) to 1.0. For the BIM, PGD, SDA, and MI-FGSM algorithms, we uniformly set the maximum number of iterations to 40. Specifically, considering the similar framework shared by the BIM and PGD algorithms, we follow the computational method from the \cite{kurakin2017adversarial} and set the $\epsilon$ parameter for both to 0.002. Through experimentation, we find that when the $\epsilon$ parameter of MI-FGSM is set to 0.01, its attack performance is the most pronounced. The SLIDE algorithm, as an adversarial attack algorithm targeting the $l_1$ norm, differs from other algorithms like PGD in terms of attack strategy. To match the performance of the SLIDE algorithm with that of PGD, we set the $l_1$ norm parameter in the AdverTorch framework to 10,000 and set the $l_1$ sparsity parameter to 0.95. For ML-DP, ML-CW, and MLA-LP, we adopt the same settings as in the original papers. 

For all algorithms except MLA-LP, we select 1000 samples that are correctly predicted by the model from each of the four datasets as attack samples. If there are fewer than 1000 correctly predicted samples, we select all of the correctly predicted samples for the attack. Regarding MLA-LP, due to limited computational resources, we take 200 samples from each of the four datasets to serve as adversarial examples for the attack.

\subsection{Results Analyses}

Our objective is to evaluate the performance of nine attack algorithms in the "Showing Many Labels" across two models and four datasets. In the experiments, following the method described in Section \ref{section 3}, we respectively calculate the $avgLabels$ for samples that extracted as attack samples from four datasets , with VOC2007 at 1.293, VOC2012 at 1.207, NUS-WIDE at 1.351, and COCO at 1.641. Subsequently, we set a series of $expLabels$ for each dataset based on different values of $n$. Specifically, VOC2007 and VOC2012 have six distinct $expLabels$, while NUS-WIDE and COCO have eight distinct $expLabels$. We record the attack success rate of each attack algorithm under eight different scenarios targeting various $expLabels$. The results are shown in Table \ref{tab2} and \ref{tab3}, where boldface indicates the best results. Since the total number of labels may vary across different datasets, we use the value of $n$ to represent different $expLabels$, with the calculation method following Equation \eqref{eq3}.
\begin{table*}
    \centering
    \caption{Evaluation Results on VOC2007 and VOC2012}
    \begin{tabular}{c|c|c|cccccc}
         \toprule[2pt]
         model & dataset & attack method & $n=0$ & $n=1$ & $n=2$ & $n=3$ & $n=4$ & $n=5$ \\
         \midrule
         \multirow{18}{*}{ML-GCN} & \multirow{9}{*}{VOC2007} & FGSM & 0.009 & 0.000 & 0.000 & 0.000 & 0.000 & 0.000 \\
         & & FGM & 0.481 & 0.138 & 0.000 & 0.000 & 0.000 & 0.000 \\
         & & BIM & \textbf{1.000} & \textbf{1.000} & \textbf{1.000} & \textbf{1.000} & \textbf{1.000} & 0.989 \\
         & & PGD & \textbf{1.000} & \textbf{1.000} & \textbf{1.000} & \textbf{1.000} & \textbf{1.000} & \textbf{1.000} \\
         & & SLIDE & \textbf{1.000} & \textbf{1.000} & \textbf{1.000} & \textbf{1.000} & 0.998 & 0.924 \\
         & & MI-FGSM & \textbf{1.000} & \textbf{1.000} & \textbf{1.000} & \textbf{1.000} & \textbf{1.000} & \textbf{1.000} \\
         & & ML-DP & 0.469 & 0.029 & 0.007 & 0.005 & 0.000 & 0.000 \\
         & & ML-CW & \textbf{1.000} & \textbf{1.000} & \textbf{1.000} & \textbf{1.000} & \textbf{1.000} & 0.993 \\
         & & MLA-LP & 0.670 & 0.225 & 0.005 & 0.000 & 0.000 & 0.000 \\
         \cline{2-9}
         &\multirow{9}{*}{VOC2012} & FGSM & 0.285 & 0.078 & 0.001 & 0.000 & 0.000 & 0.000 \\
         & & FGM & 0.415 & 0.156 & 0.008 & 0.000 & 0.000 & 0.000 \\
         & & BIM & \textbf{1.000} & \textbf{1.000} & \textbf{1.000} & \textbf{1.000} & \textbf{1.000} & \textbf{1.000} \\
         & & PGD & \textbf{1.000} & \textbf{1.000} & \textbf{1.000} & \textbf{1.000} & \textbf{1.000} & \textbf{1.000} \\
         & & SLIDE & \textbf{1.000} & \textbf{1.000} & \textbf{1.000} & \textbf{1.000} & 0.996 & 0.962 \\
         & & MI-FGSM & \textbf{1.000} & \textbf{1.000} & \textbf{1.000} & \textbf{1.000} & \textbf{1.000} & \textbf{1.000} \\
         & & ML-DP & 0.187 & 0.020 & 0.000 & 0.000 & 0.000 & 0.000 \\
         & & ML-CW & \textbf{1.000} & \textbf{1.000} & \textbf{1.000} & \textbf{1.000} & \textbf{1.000} & 0.994 \\
         & & MLA-LP & 0.660 & 0.320 & 0.050 & 0.000 & 0.000 & 0.000 \\
         \midrule
         \multirow{18}{*}{ML-LIW} & \multirow{9}{*}{VOC2007} & FGSM & 0.293 & 0.041 & 0.000 & 0.000 & 0.000 & 0.000 \\
         & & FGM & 0.414 & 0.066 & 0.000 & 0.000 & 0.000 & 0.000 \\
         & & BIM & 0.967 & 0.944 & 0.841 & 0.303 & 0.001 & 0.000 \\
         & & PGD & 0.980 & 0.949 & 0.858 & 0.299 & 0.000 & 0.000 \\
         & & SLIDE & 0.847 & 0.726 & 0.417 & 0.043 & 0.000 & 0.000 \\
         & & MI-FGSM & 0.993 & 0.989 & 0.976 & 0.920 & 0.159 & 0.014 \\
         & & ML-DP & 0.777 & 0.043 & 0.000 & 0.000 & 0.000 & 0.000 \\
         & & ML-CW & \textbf{1.000} & \textbf{1.000} & \textbf{1.000} & \textbf{1.000} & \textbf{0.662} & \textbf{0.448} \\
         & & MLA-LP & 0.35 & 0.015 & 0.000 & 0.000 & 0.000 & 0.000 \\
         \cline{2-9}
         &\multirow{9}{*}{VOC2012} & FGSM & 0.417 & 0.133 & 0.009 & 0.000 & 0.000 & 0.000 \\
         & & FGM & 0.388 & 0.114 & 0.008 & 0.000 & 0.000 & 0.000 \\
         & & BIM & 0.999 & 0.999 & 0.996 & 0.988 & 0.829 & 0.526 \\
         & & PGD & \textbf{1.000} & \textbf{1.000} & 0.997 & 0.991 & 0.843 & 0.533 \\
         & & SLIDE & 0.989 & 0.982 & 0.965 & 0.881 & 0.378 & 0.121 \\
         & & MI-FGSM & \textbf{1.000} & \textbf{1.000} & \textbf{1.000} & \textbf{1.000} & \textbf{0.996} & \textbf{0.973} \\
         & & ML-DP & 0.865 & 0.214 & 0.000 & 0.000 & 0.000 & 0.000 \\
         & & ML-CW & \textbf{1.000} & \textbf{1.000} & \textbf{1.000} & \textbf{1.000} & 0.965 & 0.422 \\
         & & MLA-LP & 0.239 & 0.034 & 0.000 & 0.000 & 0.000 & 0.000 \\
         \bottomrule[2pt]
    \end{tabular}
    \label{tab2}
\end{table*}

\begin{table*}
    \centering
    \setlength{\tabcolsep}{1mm}
    \caption{Evaluation Results on NUS-WIDE and COCO}
    \resizebox{\textwidth}{!}{
    \begin{tabular}{c|c|c|cccccccc}
         \toprule[2pt]
         model & dataset & attack method & $n=0$ & $n=1$ & $n=2$ & $n=3$ & $n=4$ & $n=5$ & $n=6$ & $n=7$ \\
         \midrule
         \multirow{18}{*}{ML-GCN} & \multirow{9}{*}{NUS-WIDE} & FGSM & 0.003 & 0.000 & 0.000 & 0.000 & 0.000 & 0.000 & 0.000 & 0.000 \\
         & & FGM & 0.267 & 0.062 & 0.007 & 0.000 & 0.000 & 0.000 & 0.000 & 0.000 \\
         & & BIM & \textbf{1.000} & \textbf{1.000} & \textbf{1.000} & 0.998 & 0.980 & 0.843 & 0.139 & 0.000 \\
         & & PGD & \textbf{1.000} & \textbf{1.000} & \textbf{1.000} & \textbf{1.000} & 0.990 & 0.878 & 0.098 & 0.000 \\
         & & SLIDE & \textbf{1.000} & \textbf{1.000} & 0.994 & 0.972 & 0.883 & 0.407 & 0.001 & 0.000 \\
         & & MI-FGSM & \textbf{1.000} & \textbf{1.000} & \textbf{1.000} & \textbf{1.000} & \textbf{1.000} & \textbf{1.000} & 0.995 & 0.832 \\
         & & ML-DP & 0.989 & 0.729 & 0.003 & 0.000 & 0.000 & 0.000 & 0.000 & 0.000 \\
         & & ML-CW & \textbf{1.000} & \textbf{1.000} & \textbf{1.000} & \textbf{1.000} & \textbf{1.000} & \textbf{1.000} & \textbf{1.000} & \textbf{0.992} \\
         & & MLA-LP & 0.590 & 0.230 & 0.055 & 0.000 & 0.000 & 0.000 & 0.000 & 0.000 \\
         \cline{2-11}
         &\multirow{9}{*}{COCO} & FGSM & 0.044 & 0.001 & 0.000 & 0.000 & 0.000 & 0.000 & 0.000 & 0.000 \\
         & & FGM & 0.533 & 0.155 & 0.008 & 0.000 & 0.000 & 0.000 & 0.000 & 0.000 \\
         & & BIM & \textbf{1.000} & \textbf{1.000} & \textbf{1.000} & \textbf{1.000} & \textbf{1.000} & \textbf{1.000} & 0.963 & 0.632 \\
         & & PGD & \textbf{1.000} & \textbf{1.000} & \textbf{1.000} & \textbf{1.000} & \textbf{1.000} & \textbf{1.000} & 0.988 & 0.683 \\
         & & SLIDE & \textbf{1.000} & \textbf{1.000} & \textbf{1.000} & \textbf{1.000} & 0.997 & 0.976 & 0.702 & 0.106 \\
         & & MI-FGSM & \textbf{1.000} & \textbf{1.000} & \textbf{1.000} & \textbf{1.000} & \textbf{1.000} & \textbf{1.000} & \textbf{1.000} & \textbf{1.000} \\
         & & ML-DP & 0.513 & 0.097 & 0.004 & 0.000 & 0.000 & 0.000 & 0.000 & 0.000 \\
         & & ML-CW & \textbf{1.000} & \textbf{1.000} & \textbf{1.000} & \textbf{1.000} & \textbf{1.000} & \textbf{1.000} & \textbf{1.000} & 0.998 \\
         & & MLA-LP & 0.807 & 0.500 & 0.156 & 0.010 & 0.000 & 0.000 & 0.000 & 0.000 \\
         \midrule
         \multirow{18}{*}{ML-LIW} & \multirow{9}{*}{NUS-WIDE} & FGSM & 0.810 & 0.546 & 0.172 & 0.003 & 0.000 & 0.000 & 0.000 & 0.000 \\
         & & FGM & 0.781 & 0.488 & 0.145 & 0.002 & 0.000 & 0.000 & 0.000 & 0.000 \\
         & & BIM & \textbf{1.000} & \textbf{1.000} & 0.955 & 0.871 & 0.684 & 0.440 & 0.042 & 0.000 \\
         & & PGD & \textbf{1.000} & \textbf{1.000} & 0.989 & 0.816 & 0.573 & 0.350 & 0.020 & 0.000 \\
         & & SLIDE & \textbf{1.000} & 0.998 & 0.912 & 0.576 & 0.386 & 0.260 & 0.000 & 0.000 \\
         & & MI-FGSM & \textbf{1.000} & \textbf{1.000} & \textbf{1.000} & 0.998 & 0.938 & 0.672 & 0.245 & 0.000 \\
         & & ML-DP & \textbf{1.000} & 0.999 & 0.914 & 0.122 & 0.000 & 0.000 & 0.000 & 0.000 \\
         & & ML-CW & \textbf{1.000} & \textbf{1.000} & \textbf{1.000} & \textbf{1.000} & \textbf{1.000} & \textbf{1.000} & \textbf{1.000} & 0.000 \\
         & & MLA-LP & 0.730 & 0.420 & 0.130 & 0.020 & 0.000 & 0.000 & 0.000 & 0.000 \\
         \cline{2-11}
         &\multirow{9}{*}{COCO} & FGSM & 0.993 & 0.960 & 0.832 & 0.474 & 0.024 & 0.000 & 0.000 & 0.000 \\
         & & FGM & 0.742 & 0.492 & 0.199 & 0.043 & 0.002 & 0.000 & 0.000 & 0.000 \\
         & & BIM & \textbf{1.000} & \textbf{1.000} & \textbf{1.000} & 0.999 & 0.997 & 0.988 & 0.882 & 0.000 \\
         & & PGD & \textbf{1.000} & \textbf{1.000} & \textbf{1.000} & 0.999 & 0.999 & 0.991 & 0.881 & 0.000 \\
         & & SLIDE & \textbf{1.000} & \textbf{1.000} & 0.995 & 0.992 & 0.966 & 0.902 & 0.542 & 0.000 \\
         & & MI-FGSM & \textbf{1.000} & \textbf{1.000} & \textbf{1.000} & \textbf{1.000} & \textbf{1.000} & \textbf{1.000} & 0.996 & 0.000 \\
         & & ML-DP & 0.999 & 0.993 & 0.981 & 0.976 & 0.072 & 0.000 & 0.000 & 0.000 \\
         & & ML-CW & \textbf{1.000} & \textbf{1.000} & \textbf{1.000} & \textbf{1.000} & \textbf{1.000} & \textbf{1.000} & \textbf{1.000} & 0.000 \\
         & & MLA-LP & 0.880 & 0.630 & 0.355 & 0.135 & 0.010 & 0.000 & 0.000 & 0.000 \\
         \bottomrule[2pt]
    \end{tabular}
    }
    \label{tab3}
\end{table*}
Experimental results indicate that different attack algorithms exhibit varying capabilities in showing labels across different models and datasets. Some algorithms are only capable of showing a small number of labels, while others can show all labels. Some algorithms perform well under certain models and datasets but perform poorly in other situations.

The overall performance of one-step attacks is significantly inferior to that of iterative attacks. When $n \geq 2$, FGSM and FGM are unable to carry out attacks, with FGSM being the worst performer among all algorithms. When attacking the ML-GCN trained on the NUS-WIDE dataset, even with $n=0$, the attack success rate is only 0.3\%. The FGSM attack performs best when targeting the ML-LIW trained on the NUS-WIDE, with attack success rates of 99.3\%, 96.0\%, 83.2\%, and 47.4\% for $n=0,1,2,3$ respectively. The FGM performs more stably due to the $l_2$ norm normalization, which ensures that perturbations are applied in the direction opposite to the gradient, but the overall performance remains poor. 

When the attacked models trained on the VOC2007 and VOC2012 datasets, both BIM and PGD generally achieve satisfactory results, especially when attacking ML-GCN, where PGD can show all labels with a 100\% success rate, and BIM shows all labels with a 98.9\% success rate on VOC2007 and 100\% on VOC2012. When the attacked models trained on the NUS-WIDE and COCO datasets, showing all labels is only possible when the dataset is COCO and the model is ML-GCN, with success rates of 63.2\% and 68.3\%, respectively. Since the NUS-WIDE and COCO datasets contain 81 and 80 categories, respectively, showing all labels on these two datasets is much more difficult than on VOC2007 and VOC2012. BIM and PGD perform worst when the model is ML-LIW and the dataset is NUS-WIDE, but in other cases, they can make samples show more than $\lfloor avgLabels+2^5 \rfloor$ labels with a high probability.

SLIDE is an $l_1$-norm variant of PGD. Except ML-GCN trained on VOC2007 and VOC2012, where its performance is comparable to PGD, in other scenarios its attack performance is significantly inferior to that of PGD. This may be because SLIDE controls the magnitude of the perturbation by limiting the $l_1$ norm, resulting in perturbations generated by SLIDE that are much smaller than those of PGD, thereby leading to a decrease in attack performance. 

The overall attack effectiveness of MI-FGSM and ML-CW is superior to other attack algorithms, especially when attacking ML-GCN trained on the NUS-WIDE dataset, where other algorithms fail to show all labels, while MI-FGSM and ML-CW achieve this with success rates of 83.2\% and 99.2\%, respectively. 

The performance of ML-DP is slightly better only when attacking the ML-LIW trained on the COCO dataset, where it can achieve an attack success rate of 97.6\% when the $n=3$. Under other conditions, once the $n \geq 2$, the attack success rate of ML-DP is generally very low. 

Among the iterative attack algorithms, MLA-LP exhibits the poorest attack performance. MLA-LP can only achieve a high success rate when the $n=0$, with the highest rate reaching up to 88\%. As the $n$ increases, its success rate drops rapidly. Once the $n \geq 2$, the algorithm is unable to carry out the attack.

\section{Conclusion}
\label{section 6}

This paper investigates the performance of attack algorithms under eight different conditions within the new attack type "Showing Many Labels", and presents the attack success rates of various attack algorithms for different $expLabels$. We utilize two multi-label classification models, four multi-label datasets, and nine attack algorithms, eight of which are adapted from multi-class environments to the multi-label environment. Experimental results indicate that under the "Showing Many Labels" attacks, the performance of one-step attacks is significantly inferior to that of iterative attacks. Furthermore, as the $expLabels$ increases, the attack becomes more challenging. However, in certain scenarios, there are still attack algorithms capable of showing all labels in the dataset with a high probability of success. In the future, we will conduct further research on more effective attack methods targeting the multi-label learning domain, as well as corresponding defensive measures.

\section*{Acknowledgements}
This study is supported by University Stability Support Program of Shenzhen (Grant No. GXWD20231130113127003), the Major Key Project of PCL (Grant No. PCL2022A03), Shenzhen Science and Technology Program (Grant No. ZDSY
S20210623091809029), Guangdong Provincial Key Laboratory of Novel Security Intelligence Technologies (Grant No. 2022B1212010005).

%
%
%
\bibliographystyle{splncs04}
%
\bibliography{samplepaper}
\end{document}